\documentclass{article}
\usepackage{ijcai16}
\usepackage{times}
\usepackage{epsfig}
\usepackage{graphicx}
\usepackage{subfigure}
\usepackage{amsmath}
\usepackage{amssymb}
\usepackage{array}
\usepackage{algorithm}
\usepackage{algorithmic}
\usepackage{makecell}
\usepackage{times}

\def\b{{\bf b}}
\def\bbb{{\bf b}}

\def\u{{\bf u}}
\def\v{{\bf v}}

\def\x{{\bf x}}

\def\z{{\bf z}}

\def\W{{\bf W}}

\def\Z{{\bf Z}}

\def\0{{\bf 0}}
\def\1{{\bf 1}}
\def\2{{\bf 2}}
\def\3{{\bf 3}}
\def\4{{\bf 4}}
\def\5{{\bf 5}}
\def\6{{\bf 6}}
\def\7{{\bf 7}}
\def\8{{\bf 8}}
\def\9{{\bf 9}}

\def\BM{{\mathcal B}}

\def\SM{{\mathcal S}}

\def\UM{{\mathcal U}}

\def\XM{{\mathcal X}}

\def\RB{{\mathbb R}}

\title{Feature Learning based Deep Supervised Hashing with Pairwise Labels}
\author{Wu-Jun Li, Sheng Wang \and Wang-Cheng Kang\\
National Key Laboratory for Novel Software Technology \\
Department of Computer Science and Technology, Nanjing University, China \\
\texttt{liwujun@nju.edu.cn, wangs@lamda.nju.edu.cn, kwc.oliver@gmail.com}}

\begin{document}
\maketitle

\begin{abstract}
  Recent years have witnessed wide application of hashing for large-scale image retrieval. However, most existing hashing methods are based on hand-crafted features which might not be optimally compatible with the hashing procedure. Recently, deep hashing methods have been proposed to perform simultaneous feature learning and hash-code learning with deep neural networks, which have shown better performance than traditional hashing methods with hand-crafted features. Most of these deep hashing methods are supervised whose supervised information is given with triplet labels. For another common application scenario with pairwise labels, there have not existed methods for simultaneous feature learning and hash-code learning. In this paper, we propose a novel deep hashing method, called deep pairwise-supervised hashing~(DPSH), to perform simultaneous feature learning and hash-code learning for applications with pairwise labels. Experiments on real datasets show that our DPSH method can outperform other methods to achieve the state-of-the-art performance in image retrieval applications.

\end{abstract}

\section{Introduction}\label{sec:intro}

With the explosive growing of data in real applications like image retrieval, approximate nearest neighbor~(ANN) search~\cite{andoni:near-optimal} has become a hot research topic in recent years. Among existing ANN techniques, hashing has become one of the most popular and effective techniques due to its fast query speed and low memory cost~\cite{kulis:kernelized,gong:procrustean,kong:isotropic,liu:kernels,rastegari:dual-view,DBLP:conf/cvpr/HeWS13,lin2014fast,DBLP:conf/cvpr/ShenSLS15,DBLP:conf/aaai/KangLZ16}.

Existing hashing methods can be divided into data-independent methods and data-dependent methods~\cite{gong:procrustean,kong:isotropic}. In data-independent methods, the hash function is typically randomly generated which is independent of any training data. The representative data-independent methods include locality-sensitive hashing~(LSH)~\cite{andoni:near-optimal} and its variants. Data-dependent methods try to learn the hash function from some training data, which is also called learning to hash~(L2H) methods~\cite{kong:isotropic}. Compared with data-independent methods, L2H methods can achieve comparable or better accuracy with shorter hash codes. Hence, L2H methods have become more and more popular than data-independent methods in real applications.

The L2H methods can be further divided into two categories~\cite{kong:isotropic,DBLP:conf/aaai/KangLZ16}: unsupervised methods and supervised methods. Unsupervised methods only utilize the feature~(attribute) information of data points without using any supervised~(label) information during the training procedure. Representative unsupervised methods include iterative quantization~(ITQ)~\cite{gong:procrustean}, isotropic hashing~(IsoHash)~\cite{kong:isotropic}, discrete graph hashing~(DGH)~\cite{DBLP:conf/nips/LiuMKC14}, and scalable graph hashing~(SGH)~\cite{DBLP:conf/ijcai/JiangL15}. Supervised methods try to utilize supervised~(label) information to learn the hash codes. The supervised information can be given in three different forms: \emph{point-wise labels}, \emph{pairwise labels} and \emph{ranking labels}. Representative point-wise label based methods include CCA-ITQ~\cite{gong:procrustean}, supervised discrete hashing~(SDH)~\cite{DBLP:conf/cvpr/ShenSLS15} and the deep hashing method in~\cite{DBLP:conf/cvpr/LinYHC15}. Representative pairwise label based methods include sequential projection learning for hashing~(SPLH)~\cite{wang:sequential},
minimal loss hashing~(MLH)~\cite{norouzi:minimal}, supervised hashing with kernels~(KSH)~\cite{liu:kernels}, two-step hashing~(TSH)~\cite{lin2013general}, fast supervised hashing~(FastH)~\cite{lin2014fast}, latent factor hashing~(LFH)~\cite{zhang2014supervised}, convolutional neural network hashing~(CNNH)~\cite{DBLP:conf/aaai/XiaPLLY14}, and column sampling based discrete supervised hashing~(COSDISH)~\cite{DBLP:conf/aaai/KangLZ16}. Representative ranking label based methods include ranking-based supervised hashing~(RSH)~\cite{DBLP:conf/iccv/WangLSJ13}, column generation hashing~(CGHash)~\cite{li:column}, order preserving hashing~(OPH)~\cite{DBLP:conf/mm/WangWYL13}, ranking preserving hashing~(RPH)~\cite{DBLP:conf/ijcai/WangZS15}, and some deep hashing methods~\cite{DBLP:conf/cvpr/ZhaoHWT15,DBLP:conf/cvpr/LaiPLY15,DBLP:journals/tip/ZhangLZZZ15}

Although a lot of hashing methods have been proposed as shown above, most existing hashing methods, including some deep hashing methods~\cite{salakhutdinov:semantic,DBLP:conf/iclr/MasciBBSS14,DBLP:conf/cvpr/LiongLWMZ15}, are based on hand-crafted features. In these methods, the hand-crafted feature construction procedure is independent of the hash-code and hash function learning procedure, and then the resulted features might not be optimally compatible with the hashing procedure. Hence, these existing hand-crafted feature based hashing methods might not achieve satisfactory performance in practice. To overcome the shortcoming of existing hand-crafted feature based methods, some feature learning based deep hashing methods~\cite{DBLP:conf/cvpr/ZhaoHWT15,DBLP:conf/cvpr/LaiPLY15,DBLP:journals/tip/ZhangLZZZ15} have recently been proposed to perform simultaneous feature learning and hash-code learning with deep neural networks, which have shown better performance than traditional hashing methods with hand-crafted features. Most of these deep hashing methods are supervised whose supervised information is given with triplet labels which are a special case of ranking labels.

For another common application scenario with pairwise labels, there have appeared few feature learning based deep hashing methods. To the best of our knowledge, \mbox{CNNH}~\cite{DBLP:conf/aaai/XiaPLLY14} is the only one which adopts deep neural network, which is actually a convolutional neural network~(\mbox{CNN})~\cite{DBLP:journals/neco/LeCunBDHHHJ89}, to perform feature learning for supervised hashing with pairwise labels. CNNH is a two-stage method. In the first stage, the hash codes are learned from the pairwise labels, and then the second stage tries to learn the hash function and feature representation from image pixels based on the hash codes from the first stage. In CNNH, the learned feature representation in the second stage cannot give feedback for learning better hash codes in the first stage. Hence, CNNH cannot perform simultaneous feature learning and hash-code learning, which still has limitations. This has been verified by the authors of CNNH themselves in another paper~\cite{DBLP:conf/cvpr/LaiPLY15}.

In this paper, we propose a novel deep hashing method, called \emph{deep pairwise-supervised hashing}~(DPSH), for applications with pairwise labels. The main contributions of DPSH are outlined as follows:
\begin{itemize}
\item DPSH is an end-to-end learning framework which contains three key components. The first component is a deep neural network to learn image representation from pixels. The second component is a hash function to map the learned image representation to hash codes. And the third component is a loss function to measure the quality of hash codes guided by the pairwise labels. All the three components are seamlessly integrated into the same deep architecture to map the images from pixels to the pairwise labels in an end-to-end way. Hence, different components can give feedback to each other in DPSH, which results in learning better codes than other methods without end-to-end architecture.
\item To the best of our knowledge, DPSH is the first method which can perform simultaneous feature learning and hash-code learning for applications with pairwise labels.
\item Experiments on real datasets show that DPSH can outperform other methods to achieve the state-of-the-art performance in image retrieval applications.
\end{itemize}


\section{Notation and Problem Definition}\label{sec:problem}

\subsection{Notation}
We use boldface lowercase letters like $\z$ to denote vectors. Boldface uppercase letters like
$\Z$ are used to denote matrices. The transpose of $\Z$ is denoted as $\Z^T$. $\|\cdot\|_2$ is used to denote the Euclidean norm of a vector. $sgn(\cdot)$ denotes the element-wise sign function which returns 1 if the element is positive and returns -1 otherwise.

\subsection{Problem Definition}
Suppose we have $n$ points~(images) $\XM=\{\x_i\}_{i=1}^n$ where $\x_i$ is the feature vector of point $i$. $\x_i$ can be the hand-crafted features or the raw pixels in image retrieval applications. The specific meaning of $\x_i$ can be easily determined from the context. Besides the feature vectors, the training set of supervised hashing with pairwise labels also contains a set of pairwise labels $\SM = \{s_{ij}\}$ with $s_{ij} \in \{0,1\}$, where $s_{ij} = 1$ means that $\x_i$ and $\x_j$ are similar, $s_{ij} = 0$ means that  $\x_i$ and $\x_j$ are dissimilar. Here, the pairwise labels typically refer to semantic labels provided with manual effort.

The goal of supervised hashing with pairwise labels is to learn a binary code $\bbb_i\in \{-1,1\}^c$ for each point $\x_i$, where $c$ is the code length. The binary codes $\BM = \{\bbb_i\}_{i=1}^n$ should preserve the similarity in $\SM$. More specifically, if $s_{ij} =1$, the binary codes $\bbb_i$ and $\bbb_j$ should have a low Hamming distance. Otherwise if $s_{ij} =0$, the binary codes $\bbb_i$ and $\bbb_j$ should have a high Hamming distance. In general, we can write the binary code as $\bbb_i = h(\x_i) = [h_1(\x_i),h_2(\x_i),\cdots,h_c(\x_i)]^T$, where $h(\x_i)$ is the hash function to learn.

\section{Model and Learning}\label{sec:model}
Most existing pairwise label based supervised hashing methods, including SPLH~\cite{wang:sequential},
MLH~\cite{norouzi:minimal}, KSH~\cite{liu:kernels}, TSH~\cite{lin2013general}, FastH~\cite{lin2014fast}, and LFH~\cite{zhang2014supervised}, adopt hand-crafted features for hash function learning. As stated in Section~\ref{sec:intro}, these methods cannot achieve satisfactory performance because the hand-crafted features might not be optimally compatible with the hash function learning procedure. CNNH~\cite{DBLP:conf/aaai/XiaPLLY14} adopts CNN to perform feature learning from raw pixels. However, CNNH is a two-stage method which cannot perform simultaneous feature learning and hash-code learning in an end-to-end way.

In this section, we introduce our model, called \emph{deep pairwise-supervised hashing}~(DPSH), which can perform simultaneous feature learning and hash-code learning in an end-to-end framework.
\begin{figure}[tb]
     \centering
  \includegraphics[width=3.3in,height = 1.65in]{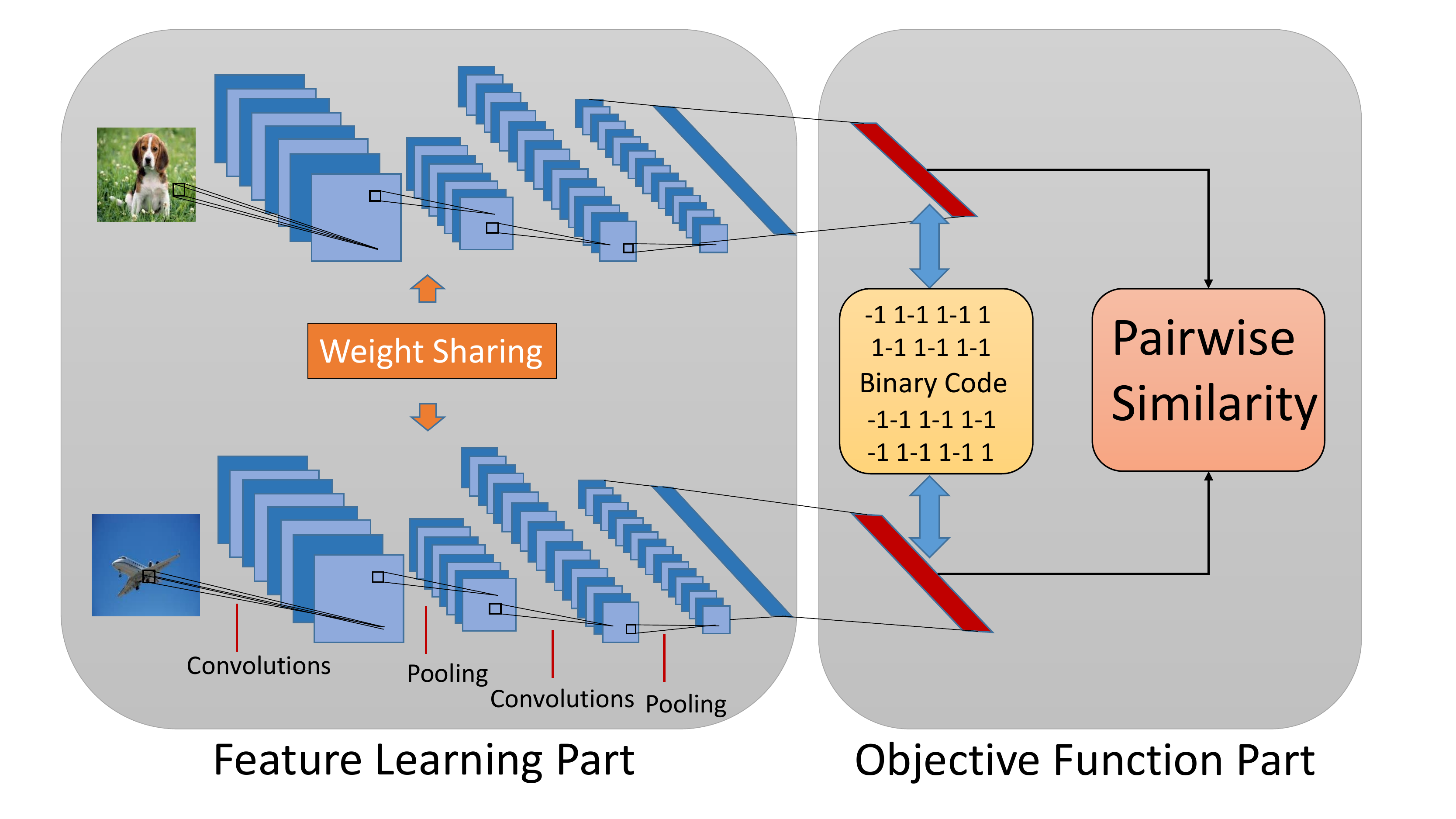}
  \vspace{-0.5cm}
  \caption{\small The end-to-end deep learning architecture for DPSH.}\label{fig:Structure}
\end{figure}

\subsection{Model}
Figure~\ref{fig:Structure} shows the end-to-end deep learning architecture for our DPSH model, which contains the feature learning part and the objective function part.

\subsubsection{Feature Learning Part}
Our DPSH model contains a CNN model from~\cite{Chatfield14} as a component. More specifically, the feature learning part has seven layers which are the same as those of CNN-F in~\cite{Chatfield14}. Other CNN architectures, such as the AlexNet~\cite{DBLP:conf/nips/KrizhevskySH12}, can also be used to substitute the CNN-F network in our DPSH model. But it is not the focus of this paper to study different networks. Hence, we just use \mbox{CNN-F} for illustrating the effectiveness of our DPSH model, and leave the study of other candidate networks for future pursuit. Please note that there are two CNNs~(top CNN and bottom CNN) in Figure~\ref{fig:Structure}. These two CNNs have the same structure and share the same weights. That is to say, both the input and loss function are based on pairs of images.

The detailed configuration of the feature learning part of our DPSH model is shown in Table~\ref{table:configuration}. More specifically, it contains 5 convolutional layers (conv 1-5) and 2 fully-connected layers (full 6-7). Each convolutional layer is described in several aspects: ``filter" specifies the number of convolution filters and their receptive field size, denoted as ``num x size x size"; ``stride" indicates the convolution stride which is the interval at which to apply the filters to the input; ``pad" indicates the number of pixels to add to each side of the input; ``\mbox{LRN}" indicates whether Local Response Normalization~(\mbox{LRN})~\cite{DBLP:conf/nips/KrizhevskySH12} is applied; ``pool" indicates the downsampling factor. ``4096" in the fully-connected layer indicates the dimensionality of the output. The activation function for all layers is the REctification Linear Unit~(RELU)~\cite{DBLP:conf/nips/KrizhevskySH12}.
\begin{table}[h]
\caption{\small Configuration of the feature learning part in DPSH. }\label{table:configuration}
  \centering
  \small
  \begin{tabular}{|c|c|}
    \hline
    Layer & Configuration \\ \hline \hline
   conv1 & filter 64x11x11, stride 4x4, pad 0, LRN, pool 2x2 \\ \hline
   conv2 & filter 256x5x5, stride 1x1, pad 2, LRN, pool 2x2 \\ \hline
   conv3 & filter 256x3x3, stride 1x1, pad 1 \\ \hline
   conv4 & filter 256x3x3, stride 1x1, pad 1 \\ \hline
   conv5 & filter 256x3x3, stride 1x1, pad 1, pool 2x2 \\ \hline
   full6 & 4096 \\ \hline
   full7 & 4096 \\ \hline
  \end{tabular}
\end{table}

\subsubsection{Objective Function Part}

Given the binary codes $\BM =\{\b_i\}_{i=1}^n$ for all the points, we can define the likelihood of the pairwise labels $\SM=\{s_{ij}\}$ as that of LFH~\cite{zhang2014supervised}:
    \begin{equation}
    p(s_{ij}|\BM)=\left\{
    \begin{aligned}
    &\sigma(\Omega_{ij}),&s_{ij} = 1 \\
    &1-\sigma(\Omega_{ij}), ~~&s_{ij} = 0\\
    \end{aligned}
    \right.\nonumber
    \end{equation}
    where $\Omega_{ij} = \frac{1}{2}\b_{i}^{T}\b_{j}$, and $\sigma(\Omega_{ij})=\frac{1}{1+e^{-\Omega_{ij}}}$. Please note that $\b_i \in \{-1,1\}^{c}$.

By taking the negative log-likelihood of the observed pairwise labels in $\SM$, we can get the following optimization problem:
    \begin{align}\label{eq:loss}
    \min_{\BM} \mathcal{J}_1 &= - \log p(\SM|\BM)= - \sum_{s_{ij}\in\SM} {\log p(s_{ij}|\BM)}\nonumber \\
      &= - \sum_{s_{ij}\in\SM}(s_{ij}\Omega_{ij}-\log(1+e^{\Omega_{ij}})).
    \end{align}
It is easy to find that the above optimization problem can make the Hamming distance between two similar points as small as possible, and simultaneously make the Hamming distance between two dissimilar points as large as possible. This exactly matches the goal of supervised hashing with pairwise labels.

The problem in~(\ref{eq:loss}) is a discrete optimization problem, which is hard to solve. LFH~\cite{zhang2014supervised} solves it by directly relaxing $\{\b_i\}$ from discrete to continuous, which might not achieve satisfactory performance~\cite{DBLP:conf/aaai/KangLZ16}.

In this paper, we design a novel strategy which can solve the problem in~(\ref{eq:loss}) in a discrete way. First, we reformulate the problem in~(\ref{eq:loss}) as the following equivalent one:
  \begin{align}\label{eq:lossU}
    \min_{\BM, \UM} \mathcal{J}_2 &= - \sum_{s_{ij}\in\SM}(s_{ij}\Theta_{ij}-\log(1+e^{\Theta_{ij}})) \\
    s.t. \hspace{0.5cm} &\u_i = \b_i, \hspace{0.5cm} \forall i=1,2,\cdots,n  \nonumber \\
                        & \u_i \in \RB^{c\times 1}, \hspace{0.5cm} \forall i=1,2,\cdots,n \nonumber \\
                        & \b_i \in \{-1,1\}^{c}, \hspace{0.5cm} \forall i=1,2,\cdots,n \nonumber
    \end{align}
where $\Theta_{ij} = \frac{1}{2}\u_i^T\u_j$, and $\UM = \{\u_i\}_{i=1}^n$.

To optimize the problem in~(\ref{eq:lossU}), we can optimize the following regularized problem by moving the equality constraints in~(\ref{eq:lossU}) to the regularization terms:
    \begin{align}
    \min_{\BM,\UM} \mathcal{J}_3 = &- \sum_{s_{ij}\in\SM}(s_{ij}\Theta_{ij}-\log(1+e^{\Theta_{ij}})) \nonumber \\
    &+\eta \sum_{i=1}^n \|\b_i-\u_i\|_{2}^{2}, \nonumber
    \end{align}
where $\eta$ is the regularization term~(hyper-parameter).

\subsubsection{DPSH Model}
To integrate the above feature learning part and objective function part into an end-to-end framework, we set
\begin{align}
\u_{i} =  \W^T \phi(\x_{i};\theta)+\v, \nonumber
\end{align}
where $\theta$ denotes all the parameters of the seven layers in the feature learning part, $\phi(\x_{i};\theta)$ denotes the output of the \mbox{full7} layer associated with point $\x_i$, $\W \in \RB^{4096\times c}$ denotes a weight matrix, $\v\in\RB^{c\times 1}$ is a bias vector. It means that we connect the feature learning part and the objective function part into the same framework by a fully-connected layer, with the weight matrix $\W$ and bias vector $\v$. After connecting the two parts, the problem for learning becomes:

    \begin{align}\label{eq:lossReg}
    \min_{\BM,\W,\v,\theta} \mathcal{J} = &- \sum_{s_{ij}\in\SM}(s_{ij}\Theta_{ij}-\log(1+e^{\Theta_{ij}})) \\
    &+\eta \sum_{i=1}^n \|\b_i-(\W^T \phi(\x_{i};\theta)+\v)\|_{2}^{2}. \nonumber
    \end{align}

As a result, we get an end-to-end deep hashing model, called DPSH, to perform simultaneous feature learning and hash-code learning in the same framework.


\subsection{Learning}
In the DPSH model, the parameters for learning contain $\W$, $\v$, $\theta$ and $\BM$. We adopt a minibatch-based strategy for learning. More specifically, in each iteration we sample a minibatch of points from the whole training set, and then perform learning based on these sampled points.

We design an alternating method for learning. That is to say, we optimize one parameter with other parameters fixed.

The $\b_i$ can be directly optimized as follows:
\begin{align}\label{eq:op_bi}
\b_i = sgn(\u_i) = sgn(\W^T \phi(\x_{i};\theta)+\v).
\end{align}

For the other parameters $\W$, $\v$ and $\theta$, we use back-propagation~(BP) for learning. In particular, we can compute the derivatives of the loss function with respect to $\u_i$ as follows:
    \begin{align}
    \frac{\partial \mathcal{J}}{\partial \u_{i}}= & \frac{1}{2}\sum_{j:s_{ij}\in\SM} (a_{ij}-s_{ij})\u_{j} + \frac{1}{2}\sum_{j:s_{ji}\in\SM} (a_{ji}-s_{ji})\u_{j} \nonumber \\
    &+ 2\eta(\u_{i}-\b_{i}), \nonumber
    \end{align}
 where $a_{ij} = \sigma(\frac{1}{2}\u_i^T \u_j)$.

Then, we can update the parameters $\W$, $\v$ and $\theta$ by utilizing back propagation:
    \begin{align}
    &\frac{\partial \mathcal{J}}{\partial \W}=\phi(\x_{i};\theta)(\frac{\partial \mathcal{J}}{\partial \u_{i}})^{T},\label{eq:gradientW}\\
    &\frac{\partial \mathcal{J}}{\partial \v}=\frac{\partial \mathcal{J}}{\partial \u_i},\label{eq:gradientV}\\
    &\frac{\partial \mathcal{J}}{\phi(\x_{i};\theta)}=\W\frac{\partial \mathcal{J}}{\partial \u_{i}}.\label{eq:gradientTheta}
    \end{align}

The whole learning algorithm of DPSH is briefly summarized in Algorithm~\ref{alg:learning}.

\begin{algorithm}[t]
\caption{Learning algorithm for DPSH.}\label{alg:learning}
\begin{algorithmic}
\small
\STATE \textbf{Input}: \\
~~~~~Training images $\XM=\{\x_i\}_{i=1}^n$ and a set of pairwise labels $\SM = \{s_{ij}\}$.
\STATE \textbf{Output}: \\
~~~~~The parameters $\W$, $\v$, $\theta$ and $\BM$.\\

\textbf{Initialization}: Initialize $\theta$ with the CNN-F model; Initialize each entry of $\W$ and $\v$ by randomly sampling from a Gaussian distribution with mean 0 and variance $0.01$.
\STATE \textbf{REPEAT}\\
~~~~~Randomly sample a minibatch of points from $\XM$, and for each sampled point $\x_i$, perform the following operations:
\begin{itemize}
\item Calculate $\phi(\x_i;\theta)$ by forward propagation;
\item Compute $\u_i =  \W^T \phi(\x_{i};\theta)+\v$;
\item Compute the binary code of $\x_{i}$ with $\b_{i} = sgn(\u_{i})$.
\item Compute derivatives for point $\x_{i}$ according to~(\ref{eq:gradientW}), (\ref{eq:gradientV}) and (\ref{eq:gradientTheta});
\item Update the parameters $\W, \v, \theta$ by utilizing back propagation;
\end{itemize}
\STATE \textbf{UNTIL} a fixed number of iterations
\end{algorithmic}
\end{algorithm}

\subsection{Out-of-Sample Extension}
After we have completed the learning procedure, we can only get the hash codes for points in the training data. We still need to perform out-of-sample extension to predict the hash codes for the points which are not appeared in the training set.

The deep hashing framework of DPSH can be naturally applied for out-of-sample extension. For any point $\x_q \notin \XM$, we can predict its hash code just by forward propagation:
\begin{align}
  \b_q = h(\x_q) = sgn(\W^T \phi(\x_{q};\theta)+\v).
\end{align}

\section{Experiment}\label{sec:exp}

 All our experiments for DPSH are completed with MatConvNet~\cite{vedaldi15matconvnet} on a NVIDIA K80 GPU server. Our model can be trained at the speed of about 290 images per second with a single K80 GPU.

\subsection{Datasets and Setting}
We compare our model with several baselines on two widely used benchmark datasets: \emph{CIFAR-10} and \emph{NUS-WIDE}.

The CIFAR-10~\cite{krizhevsky:multiple-layers} dataset consists of 60,000 32$\times$32 color images which are categorized into 10 classes~(6000 images per class). It is a single-label dataset in which each image belongs to one of the ten classes.

The NUS-WIDE dataset~\cite{chua:nus-wide,DBLP:journals/spl/ZhaoLZ15} has nearly 270,000 images collected from the web. It is a multi-label dataset in which each image is annotated with one or mutiple class labels from 81 classes. Following~\cite{DBLP:conf/cvpr/LaiPLY15}, we only use the images associated with the 21 most frequent classes. For these classes, the number of images of each class is at least 5000.

We compare our method with several state-of-the-art hashing methods. These methods can be categorized into five classes:
\begin{itemize}
\item Unsupervised hashing methods with hand-crafted features, including SH~\cite{weiss:spectral} and ITQ~\cite{gong:procrustean}.
\item Supervised hashing methods with hand-crafted features, including SPLH~\cite{wang:sequential}, KSH~\cite{liu:kernels}, FastH~\cite{lin2014fast}, LFH~\cite{zhang2014supervised}, and SDH~\cite{DBLP:conf/cvpr/ShenSLS15}.
\item The above unsupervised methods and supervised methods with deep features extracted by the CNN-F of the feature learning part in our DPSH.
\item Deep hashing methods with pairwise labels, including CNNH~\cite{DBLP:conf/aaai/XiaPLLY14}.
\item Deep hashing methods with triplet labels, including network in network hashing~(NINH)~\cite{DBLP:conf/cvpr/LaiPLY15}, deep semantic ranking based hashing~(DSRH)~\cite{DBLP:conf/cvpr/ZhaoHWT15}, deep similarity comparison hashing~(DSCH)~\cite{DBLP:journals/tip/ZhangLZZZ15} and deep regularized similarity comparison hashing~(DRSCH)~\cite{DBLP:journals/tip/ZhangLZZZ15}.
\end{itemize}

For hashing methods which use hand-crafted features, we represent each image in CIFAR-10 by a 512-dimensional GIST vector. And we represent each image in NUS-WIDE by a 1134-dimensional low level feature vector, including 64-D color histogram, 144-D color correlogram, 73-D edge direction histogram, 128-D wavelet texture, 225-D block-wise color moments and 500-D SIFT features.

For deep hashing methods, we first resize all images to be 224$\times$224 pixels and then directly use the raw image pixels as input. We adopt the CNN-F network which has been pre-trained on the ImageNet dataset~\cite{DBLP:journals/corr/RussakovskyDSKSMHKKBBF14} to initialize the first seven layers of our DPSH framework. Similar initialization strategy has also been adopted by other deep hashing methods~\cite{DBLP:conf/cvpr/ZhaoHWT15}.

As most existing hashing methods, the mean average precision~(MAP) is used to measure the accuracy of our proposed method and other baselines. The hyper-parameter $\eta$ in DPSH is chosen by a validation set, which is $10$ for CIFAR-10 and $100$ for NUS-WIDE unless otherwise stated.

\subsection{Accuracy}\label{sec:expAcc}

Following~\cite{DBLP:conf/aaai/XiaPLLY14,DBLP:conf/cvpr/LaiPLY15}, we randomly select 1000 images (100 images per class) as the query set in CIFAR-10. For the unsupervised methods, we use the rest images as the training set. For the supervised methods, we randomly select 5000 images~(500 images per class) from the rest images as the training set. The pairwise label set $\SM$ is constructed based on the image class labels. That is to say, two images will be considered to be similar if they share the same class label.

In NUS-WIDE, we randomly sample 2100 query images from 21 most frequent labels~(100 images per class) by following the strategy in~\cite{DBLP:conf/aaai/XiaPLLY14,DBLP:conf/cvpr/LaiPLY15}. For supervised methods, we randomly select 500 images per class from the rest images as the training set. The pairwise label set $\SM$ is constructed based on the image class labels. That is to say, two images will be considered to be similar if they share at least one common label. For NUS-WIDE, we calculate the MAP values within the top 5000 returned neighbors.

The MAP results are reported in Table~\ref{table:MAP1}, where DPSH, DPSH0, NINH and CNNH are deep methods, and all the other methods are non-deep methods with hand-crafted features. The result of NINH, CNNH, KSH and ITQ are from~\cite{DBLP:conf/aaai/XiaPLLY14,DBLP:conf/cvpr/LaiPLY15}. Please note that the above experimental setting and evaluation metric is exactly the same as that in~\cite{DBLP:conf/aaai/XiaPLLY14,DBLP:conf/cvpr/LaiPLY15}. Hence, the comparison is reasonable. We can find that our method DPSH dramatically outperform other baselines\footnote{The accuracy of LFH in Table~\ref{table:MAP1} is much lower than that in~\cite{zhang2014supervised,DBLP:conf/aaai/KangLZ16} because less points are adopted for training in this paper. Please note that LFH is an efficient method which can be used for training large-scale supervised hashing problems. But the training efficiency is not the focus of this paper.}, including unsupervised methods, supervised methods with hand-crafted features, and deep hashing methods with feature learning.

Both DPSH and CNNH are deep hashing methods with pairwise labels. By comparing DPSH to CNNH, we can find that the model~(DPSH) with simultaneous feature learning and hash-code learning can outperform the other model~(CNNH) without simultaneous feature learning and hash-code learning.

NINH is a triplet label based method. Although NINH can perform simultaneous feature learning and hash-code learning, it is still outperformed by DPSH. More comparison with triplet label based methods will be provided in Section~\ref{sec:expRank}.

To further verify the importance of simultaneous feature learning and hash-code learning, we design a variant of DPSH, called DPSH0, which does not update the parameter of the first seven layers~(CNN-F layers) during learning. Hence, DPSH0 just uses the CNN-F for feature extraction, and then based on the extracted features to learn hash functions. The hash function learning procedure will give no feedback to the feature extraction procedure. By comparing DPSH to DPSH0, we find that DPSH can dramatically outperform DPSH0. It means that integrating feature learning and hash-code learning into the same framework in an end-to-end way can get a better solution than that without end-to-end learning.

\begin{table*}[htb]
\small
\centering
\caption{\small  Accuracy in terms of MAP. The best MAPs for each category are shown in boldface. Here, the MAP value is calculated based on the top 5000 returned neighbors for NUS-WIDE dataset.}\label{table:MAP1}
\begin{tabular}{|p{1.9cm}||p{1.2cm}<{\centering}|p{1.2cm}<{\centering}|p{1.2cm}<{\centering}|p{1.2cm}<{\centering}||p{1.2cm}<{\centering}|p{1.2cm}<{\centering}|p{1.2cm}<{\centering}|p{1.2cm}<{\centering}|}
\hline
Method              & \multicolumn{4}{c||}{CIFAR-10~(MAP)}     & \multicolumn{4}{c|}{NUS-WIDE~(MAP)}                                                                                                                                                                                                                                                                                                     \\ \hline \hline
                      & 12-bits                & 24-bits              & 32-bits               & 48-bits    & 12-bits                & 24-bits              & 32-bits               & 48-bits                 \\ \hline
                      DPSH   & \textbf{0.713}                             & \textbf{0.727 }                          &\textbf{0.744}                            & \textbf{0.757 }         & \textbf{0.794 }                        &\textbf{0.822 }                     &\textbf{0.838}                            &\textbf{ 0.851}                                         \\ \hline
                      DPSH0             & 0.479        & 0.472       &0.470       & 0.495        & 0.747 &0.751     &0.763 &0.776\\ \hline
NINH             & 0.552        & 0.566       &0.558       & 0.581        & 0.674       &0.697     &0.713&0.715     \\ \hline
CNNH               & 0.439       & 0476      &0.472      & 0.489         & 0.611      &0.618     &0.625&0.608     \\ \hline
FastH          & 0.305        & 0.349       &0.369       & 0.384         & 0.621      &0.650    &0.665&0.687     \\ \hline
SDH                 & 0.285        & 0.329        & 0.341   &0.356   &  0.568 & 0.600 &  0.608  & 0.637                      \\ \hline
KSH                 & 0.303      &0.337       &0.346      &0.356        & 	0.556     &0.572    & 0.581         &0.588            \\ \hline
LFH                 & 0.176      & 0.231     & 0.211    &0.253           &0.571       &0.568   &0.568     & 0.585                     \\ \hline
SPLH                 & 0.171      &0.173  &0.178    & 0.184         & 0.568   &0.589    &0.597   & 0.601       \\  \hline
ITQ               & 0.162       & 0.169       &0.172      & 0.175         & 0.452       &0.468     &0.472 &0.477     \\ \hline
SH               & 0.127        & 0.128       &0.126      & 0.129        & 0.454       &0.406    &0.405&0.400    \\ \hline
\end{tabular}
%
\caption{\small  Accuracy in terms of MAP. The best MAPs for each category are shown in boldface. Here, the MAP value is calculated based on the top 5000 returned neighbors for NUS-WIDE dataset.}\label{table:MAP2}
\begin{tabular}{|p{1.9cm}||p{1.2cm}<{\centering}|p{1.2cm}<{\centering}|p{1.2cm}<{\centering}|p{1.2cm}<{\centering}||p{1.2cm}<{\centering}|p{1.2cm}<{\centering}|p{1.2cm}<{\centering}|p{1.2cm}<{\centering}|}
\hline
Method              & \multicolumn{4}{c||}{CIFAR-10~(MAP)}     & \multicolumn{4}{c|}{NUSWIDE~(MAP)}                                                                                                                                                                                                                                                                                                     \\ \hline \hline
                      & 12-bits                & 24-bits              & 32-bits               & 48-bits    & 12-bits                & 24-bits              & 32-bits               & 48-bits                 \\ \hline
                      DPSH   & \textbf{0.713}                             & \textbf{0.727 }                          &\textbf{0.744}                            & \textbf{0.757 }         & \textbf{0.794 }                        &\textbf{0.822 }                     &\textbf{0.838}                            &\textbf{ 0.851}                                         \\ \hline
FastH + \small{CNN}               & 0.553        &0.607       &0.619      &0.636      &0.779      &0.807    &0.816    &0.825    \\ \hline
SDH + \small{CNN}               & 0.478       & 0.557        & 0.584   &0.592   &0.780    &0.804    &0.815    &0.824  \\ \hline
KSH + \small{CNN}                 &0.488  & 0.539  &0.548  &0.563       &0.768 	     &0.786      &0.790       &0.799           \\ \hline
LFH + \small{CNN}                  & 0.208     & 0.242     & 0.266    &0.339      &0.695      &0.734     &0.739     &0.759                      \\ \hline
SPLH + \small{CNN}                 & 0.299      &0.330  &0.335    &0.330          &0.753    &0.775    &0.783    &0.786        \\  \hline
ITQ + \small{CNN}               & 0.237       & 0.246       &0.255      & 0.261         &0.719    &0.739    &0.747     &0.756     \\ \hline
SH + \small{CNN}               & 0.183        & 0.164       &0.161      & 0.161        & 0.621      &0.616     &0.615      &0.612   \\ \hline
\end{tabular}

\centering
\caption{\small Accuracy in terms of MAP. The best MAPs for each category are shown in boldface. Here, the MAP value is calculated based on the top 50,000 returned neighbors for NUS-WIDE dataset. }\label{table:MAP3}
\begin{tabular}{|p{1.9cm}||p{1.2cm}<{\centering}|p{1.2cm}<{\centering}|p{1.2cm}<{\centering}|p{1.2cm}<{\centering}||p{1.2cm}<{\centering}|p{1.2cm}<{\centering}|p{1.2cm}<{\centering}|p{1.2cm}<{\centering}|}
\hline
Method              & \multicolumn{4}{c||}{CIFAR-10~(MAP)}     & \multicolumn{4}{c|}{NUS-WIDE~(MAP)}                                                                                                                                                                                                                                                                                                     \\ \hline \hline
                      & 16-bits                & 24-bits              & 32-bits               & 48-bits    & 16-bits                & 24-bits              & 32-bits               & 48-bits                 \\ \hline
                      DPSH   & \textbf{0.763}                             & \textbf{0.781 }                          &\textbf{0.795}                            & \textbf{0.807 }         & \textbf{0.715 }                        &\textbf{0.722 }                     &\textbf{0.736}                         & \textbf{0.741 }                                         \\ \hline
DRSCH            &0.615         &0.622       &0.629      &0.631      &0.618      &0.622    &0.623    &0.628    \\ \hline
DSCH            &0.609         &0.613       &0.617      &0.620      &0.592      &0.597    &0.611    &0.609    \\ \hline
DSRH            &0.608         &0.611       &0.617      &0.618      &0.609      &0.618    &0.621    &0.631    \\ \hline
\end{tabular}
\end{table*}


\subsection{Comparison to Non-Deep Baselines with Deep Features}
To further verify the effectiveness of simultaneous feature learning and hash-code learning, we compare DPSH to other non-deep methods with deep features extracted by the CNN-F pre-trained on ImageNet. The results are reported in Table~\ref{table:MAP2}, where ``FastH+CNN" denotes the FastH method with deep features and other methods have similar notations. We can find that our DPSH outperforms all the other non-deep baselines with deep features.

\subsection{Comparison to Baselines with Ranking Labels}\label{sec:expRank}

Most existing deep supervised hashing methods are based on ranking labels, especially triplet labels. Although the learning procedure of these methods is based on ranking labels, the learned model can also be used for evaluation scenario with pairwise labels. In fact, most triplet label based methods adopt pairwise labels as ground truth for evaluation~\cite{DBLP:conf/cvpr/LaiPLY15,DBLP:journals/tip/ZhangLZZZ15}. In Section~\ref{sec:expAcc}, we have shown that our DPSH can outperform NINH. In this subsection, we will perform further comparison to other deep hashing methods with ranking labels (triplet labels). These methods include DSRH~\cite{DBLP:conf/cvpr/ZhaoHWT15}, DSCH~\cite{DBLP:journals/tip/ZhangLZZZ15} and DRSCH~\cite{DBLP:journals/tip/ZhangLZZZ15}.

The experimental setting in DSCH and DRSCH~\cite{DBLP:journals/tip/ZhangLZZZ15} is different from that in Section~\ref{sec:expAcc}. To perform fair comparison, we adopt the same setting as that in~\cite{DBLP:journals/tip/ZhangLZZZ15} for evaluation. More specifically, in CIFAR-10 dataset, we randomly sample 10,000 query images (1000 images per class) and use the rest as the training set. In the NUS-WIDE dataset, we randomly sample 2100 query images from 21 most frequently happened semantic labels (100 images per class), and use the rest as training samples. For NUS-WIDE, the MAP values within the top 50,000 returned neighbors are used for evaluation.

The experimental results are shown in Table~\ref{table:MAP3}. Please note that the results of DPSH in Table~\ref{table:MAP3} are different from those in Table~\ref{table:MAP1}, because the experimental settings are different. The results of DSRH, DSCH and DRSCH are directly from~\cite{DBLP:journals/tip/ZhangLZZZ15}. From Table~\ref{table:MAP3}, we can find that DPSH with pairwise labels can also dramatically outperform other baselines with triplet labels. Please note that DSRH, DSCH and DRSCH can also perform simultaneously feature learning and hash-code learning in an end-to-end framework.

\subsection{Sensitivity to Hyper-Parameter}
\begin{figure}[t]
     \centering
  \subfigure[CIFAR-10]{\includegraphics[width=1.4in,height = 1.05in]{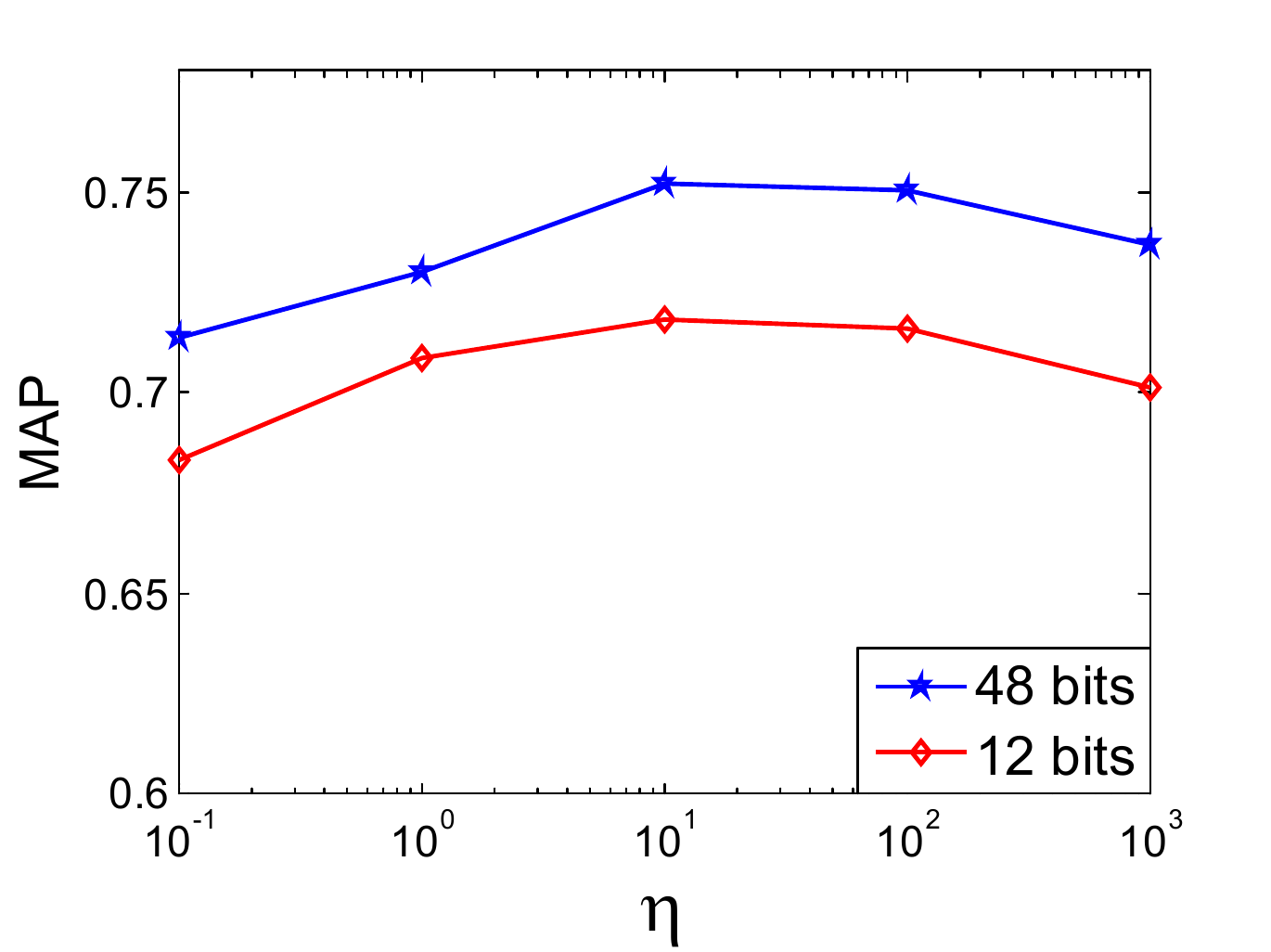}}
  \subfigure[NUS-WIDE]{\includegraphics[width=1.4in,height = 1.05in]{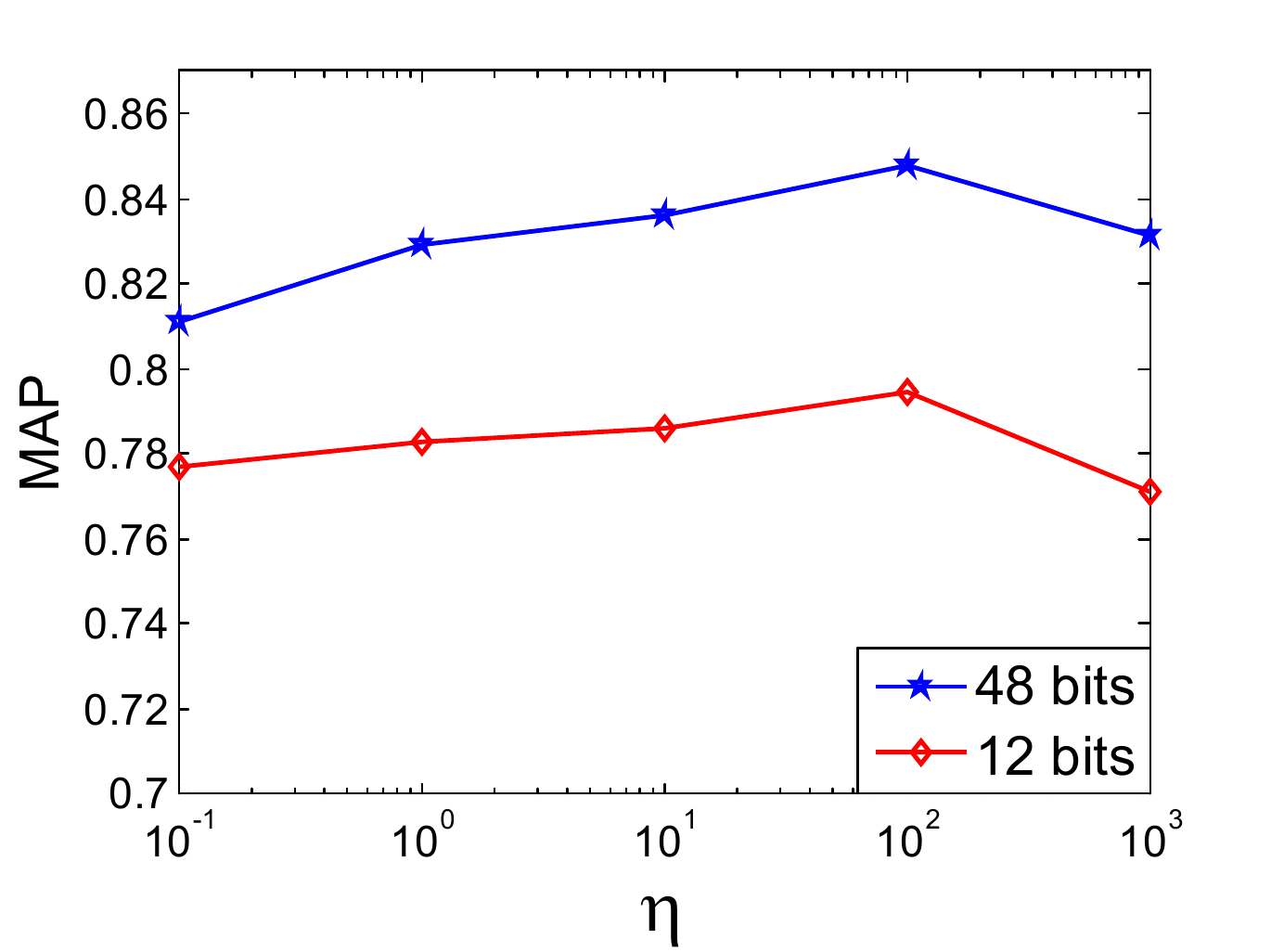}}
  \vspace{-0.4cm}
  \caption{\small Sensitivity to hyper-parameter.}\label{fig:hyperparameter}
\end{figure}

Figure~\ref{fig:hyperparameter} shows the effect of the hyper-parameter $\eta$. We can find that DPSH is not sensitive to $\eta$ in a large range. For example, DPSH can achieve good performance on both datasets with $10\leq \eta \leq 100$.

\section{Conclusion}\label{sec:conclusion}

In this paper, we have proposed a novel deep hashing methods, called DPSH, for settings with pairwise labels. To the best of our knowledge, DPSH is the first method which can perform simultaneous feature learning and hash-code learning for applications with pairwise labels. Because different components in DPSH can give feedback to each other, DPSH can learn better codes than other methods without end-to-end architecture. Experiments on real datasets show that DPSH can outperform other methods to achieve the state-of-the-art performance in image retrieval applications.


\section{Acknowledgements}
{\small This work is supported by the NSFC~(61472182), the Fundamental Research Funds for the Central Universities~(20620140510), and the Tencent Fund~(2014320001013613).}

{\small
\bibliographystyle{named}
\bibliography{DPSH_v7}

\begin{thebibliography}{}

\bibitem[\protect\citeauthoryear{Andoni and Indyk}{2006}]{andoni:near-optimal}
Alexandr Andoni and Piotr Indyk.
\newblock Near-optimal hashing algorithms for approximate nearest neighbor in
  high dimensions.
\newblock In {\em FOCS}, pages 459--468, 2006.

\bibitem[\protect\citeauthoryear{Chatfield \bgroup \em et al.\egroup
  }{2014}]{Chatfield14}
Ken Chatfield, Karen Simonyan, Andrea Vedaldi, and Andrew Zisserman.
\newblock Return of the devil in the details: Delving deep into convolutional
  nets.
\newblock In {\em BMVC}, 2014.

\bibitem[\protect\citeauthoryear{Chua \bgroup \em et al.\egroup
  }{2009}]{chua:nus-wide}
Tat-Seng Chua, Jinhui Tang, Richang Hong, Haojie Li, Zhiping Luo, and Yantao
  Zheng.
\newblock {NUS-WIDE}: A real-world web image database from national university
  of singapore.
\newblock In {\em CIVR}, 2009.

\bibitem[\protect\citeauthoryear{Gong and Lazebnik}{2011}]{gong:procrustean}
Yunchao Gong and Svetlana Lazebnik.
\newblock Iterative quantization: A procrustean approach to learning binary
  codes.
\newblock In {\em CVPR}, pages 817--824, 2011.

\bibitem[\protect\citeauthoryear{He \bgroup \em et al.\egroup
  }{2013}]{DBLP:conf/cvpr/HeWS13}
Kaiming He, Fang Wen, and Jian Sun.
\newblock K-means hashing: An affinity-preserving quantization method for
  learning binary compact codes.
\newblock In {\em CVPR}, pages 2938--2945, 2013.

\bibitem[\protect\citeauthoryear{Jiang and Li}{2015}]{DBLP:conf/ijcai/JiangL15}
Qing{-}Yuan Jiang and Wu{-}Jun Li.
\newblock Scalable graph hashing with feature transformation.
\newblock In {\em IJCAI}, pages 2248--2254, 2015.

\bibitem[\protect\citeauthoryear{Kang \bgroup \em et al.\egroup
  }{2016}]{DBLP:conf/aaai/KangLZ16}
Wang-Cheng Kang, Wu-Jun Li, and Zhi-Hua Zhou.
\newblock Column sampling based discrete supervised hashing.
\newblock In {\em AAAI}, 2016.

\bibitem[\protect\citeauthoryear{Kong and Li}{2012}]{kong:isotropic}
Weihao Kong and Wu-Jun Li.
\newblock Isotropic hashing.
\newblock In {\em NIPS}, pages 1655--1663, 2012.

\bibitem[\protect\citeauthoryear{Krizhevsky \bgroup \em et al.\egroup
  }{2012}]{DBLP:conf/nips/KrizhevskySH12}
Alex Krizhevsky, Ilya Sutskever, and Geoffrey~E. Hinton.
\newblock Imagenet classification with deep convolutional neural networks.
\newblock In {\em NIPS}, pages 1106--1114, 2012.

\bibitem[\protect\citeauthoryear{Krizhevsky}{2009}]{krizhevsky:multiple-layers}
Alex Krizhevsky.
\newblock Learning multiple layers of features from tiny images.
\newblock Master's thesis, University of Toronto, 2009.

\bibitem[\protect\citeauthoryear{Kulis and Grauman}{2009}]{kulis:kernelized}
Brian Kulis and Kristen Grauman.
\newblock Kernelized locality-sensitive hashing for scalable image search.
\newblock In {\em ICCV}, pages 2130--2137, 2009.

\bibitem[\protect\citeauthoryear{Lai \bgroup \em et al.\egroup
  }{2015}]{DBLP:conf/cvpr/LaiPLY15}
Hanjiang Lai, Yan Pan, Ye~Liu, and Shuicheng Yan.
\newblock Simultaneous feature learning and hash coding with deep neural
  networks.
\newblock In {\em CVPR}, pages 3270--3278, 2015.

\bibitem[\protect\citeauthoryear{LeCun \bgroup \em et al.\egroup
  }{1989}]{DBLP:journals/neco/LeCunBDHHHJ89}
Yann LeCun, Bernhard~E. Boser, John~S. Denker, Donnie Henderson, R.~E. Howard,
  Wayne~E. Hubbard, and Lawrence~D. Jackel.
\newblock Backpropagation applied to handwritten zip code recognition.
\newblock {\em Neural Computation}, 1(4):541--551, 1989.

\bibitem[\protect\citeauthoryear{Li \bgroup \em et al.\egroup
  }{2013}]{li:column}
Xi~Li, Guosheng Lin, Chunhua Shen, Anton van~den Hengel, and Anthony~R. Dick.
\newblock Learning hash functions using column generation.
\newblock In {\em ICML}, pages 142--150, 2013.

\bibitem[\protect\citeauthoryear{Lin \bgroup \em et al.\egroup
  }{2013}]{lin2013general}
Guosheng Lin, Chunhua Shen, David Suter, and Anton van~den Hengel.
\newblock A general two-step approach to learning-based hashing.
\newblock In {\em ICCV}, pages 2552--2559, 2013.

\bibitem[\protect\citeauthoryear{Lin \bgroup \em et al.\egroup
  }{2014}]{lin2014fast}
Guosheng Lin, Chunhua Shen, Qinfeng Shi, Anton van~den Hengel, and David Suter.
\newblock Fast supervised hashing with decision trees for high-dimensional
  data.
\newblock In {\em CVPR}, pages 1971--1978, 2014.

\bibitem[\protect\citeauthoryear{Lin \bgroup \em et al.\egroup
  }{2015}]{DBLP:conf/cvpr/LinYHC15}
Kevin Lin, Huei-Fang Yang, Jen-Hao Hsiao, and Chu-Song Chen.
\newblock Deep learning of binary hash codes for fast image retrieval.
\newblock In {\em CVPR Workshops}, pages 27--35, 2015.

\bibitem[\protect\citeauthoryear{Liong \bgroup \em et al.\egroup
  }{2015}]{DBLP:conf/cvpr/LiongLWMZ15}
Venice~Erin Liong, Jiwen Lu, Gang Wang, Pierre Moulin, and Jie Zhou.
\newblock Deep hashing for compact binary codes learning.
\newblock In {\em CVPR}, pages 2475--2483, 2015.

\bibitem[\protect\citeauthoryear{Liu \bgroup \em et al.\egroup
  }{2012}]{liu:kernels}
Wei Liu, Jun Wang, Rongrong Ji, Yu-Gang Jiang, and Shih-Fu Chang.
\newblock Supervised hashing with kernels.
\newblock In {\em CVPR}, pages 2074--2081, 2012.

\bibitem[\protect\citeauthoryear{Liu \bgroup \em et al.\egroup
  }{2014}]{DBLP:conf/nips/LiuMKC14}
Wei Liu, Cun Mu, Sanjiv Kumar, and Shih{-}Fu Chang.
\newblock Discrete graph hashing.
\newblock In {\em NIPS}, pages 3419--3427, 2014.

\bibitem[\protect\citeauthoryear{Masci \bgroup \em et al.\egroup
  }{2014}]{DBLP:conf/iclr/MasciBBSS14}
Jonathan Masci, Alex~M. Bronstein, Michael~M. Bronstein, Pablo Sprechmann, and
  Guillermo Sapiro.
\newblock Sparse similarity-preserving hashing.
\newblock In {\em ICLR}, 2014.

\bibitem[\protect\citeauthoryear{Norouzi and Fleet}{2011}]{norouzi:minimal}
Mohammad Norouzi and David~J. Fleet.
\newblock Minimal loss hashing for compact binary codes.
\newblock In {\em ICML}, pages 353--360, 2011.

\bibitem[\protect\citeauthoryear{Rastegari \bgroup \em et al.\egroup
  }{2013}]{rastegari:dual-view}
Mohammad Rastegari, Jonghyun Choi, Shobeir Fakhraei, Daume Hal, and Larry~S.
  Davis.
\newblock Predictable dual-view hashing.
\newblock In {\em ICML}, pages 1328--1336, 2013.

\bibitem[\protect\citeauthoryear{Russakovsky \bgroup \em et al.\egroup
  }{2014}]{DBLP:journals/corr/RussakovskyDSKSMHKKBBF14}
Olga Russakovsky, Jia Deng, Hao Su, Jonathan Krause, Sanjeev Satheesh, Sean Ma,
  Zhiheng Huang, Andrej Karpathy, Aditya Khosla, Michael~S. Bernstein,
  Alexander~C. Berg, and Fei{-}Fei Li.
\newblock Imagenet large scale visual recognition challenge.
\newblock {\em CoRR}, abs/1409.0575, 2014.

\bibitem[\protect\citeauthoryear{Salakhutdinov and
  Hinton}{2009}]{salakhutdinov:semantic}
Ruslan Salakhutdinov and Geoffrey~E. Hinton.
\newblock Semantic hashing.
\newblock {\em International Journal of Approximate Reasoning}, 50(7):969--978,
  2009.

\bibitem[\protect\citeauthoryear{Shen \bgroup \em et al.\egroup
  }{2015}]{DBLP:conf/cvpr/ShenSLS15}
Fumin Shen, Chunhua Shen, Wei Liu, and Heng~Tao Shen.
\newblock Supervised discrete hashing.
\newblock In {\em CVPR}, 2015.

\bibitem[\protect\citeauthoryear{Vedaldi and Lenc}{2015}]{vedaldi15matconvnet}
Andrea Vedaldi and Karel Lenc.
\newblock Mat{C}onv{N}et -- convolutional neural networks for {MATLAB}.
\newblock In {\em ACM MM}, 2015.

\bibitem[\protect\citeauthoryear{Wang \bgroup \em et al.\egroup
  }{2010}]{wang:sequential}
Jun Wang, Sanjiv Kumar, and Shih-Fu Chang.
\newblock Sequential projection learning for hashing with compact codes.
\newblock In {\em ICML}, pages 1127--1134, 2010.

\bibitem[\protect\citeauthoryear{Wang \bgroup \em et al.\egroup
  }{2013a}]{DBLP:conf/mm/WangWYL13}
Jianfeng Wang, Jingdong Wang, Nenghai Yu, and Shipeng Li.
\newblock Order preserving hashing for approximate nearest neighbor search.
\newblock In {\em {ACM} MM}, pages 133--142, 2013.

\bibitem[\protect\citeauthoryear{Wang \bgroup \em et al.\egroup
  }{2013b}]{DBLP:conf/iccv/WangLSJ13}
Jun Wang, Wei Liu, Andy~X. Sun, and Yu{-}Gang Jiang.
\newblock Learning hash codes with listwise supervision.
\newblock In {\em ICCV}, pages 3032--3039, 2013.

\bibitem[\protect\citeauthoryear{Wang \bgroup \em et al.\egroup
  }{2015}]{DBLP:conf/ijcai/WangZS15}
Qifan Wang, Zhiwei Zhang, and Luo Si.
\newblock Ranking preserving hashing for fast similarity search.
\newblock In {\em IJCAI}, pages 3911--3917, 2015.

\bibitem[\protect\citeauthoryear{Weiss \bgroup \em et al.\egroup
  }{2008}]{weiss:spectral}
Yair Weiss, Antonio Torralba, and Robert Fergus.
\newblock Spectral hashing.
\newblock In {\em NIPS}, pages 1753--1760, 2008.

\bibitem[\protect\citeauthoryear{Xia \bgroup \em et al.\egroup
  }{2014}]{DBLP:conf/aaai/XiaPLLY14}
Rongkai Xia, Yan Pan, Hanjiang Lai, Cong Liu, and Shuicheng Yan.
\newblock Supervised hashing for image retrieval via image representation
  learning.
\newblock In {\em AAAI}, pages 2156--2162, 2014.

\bibitem[\protect\citeauthoryear{Zhang \bgroup \em et al.\egroup
  }{2014}]{zhang2014supervised}
Peichao Zhang, Wei Zhang, Wu-Jun Li, and Minyi Guo.
\newblock Supervised hashing with latent factor models.
\newblock In {\em SIGIR}, pages 173--182, 2014.

\bibitem[\protect\citeauthoryear{Zhang \bgroup \em et al.\egroup
  }{2015}]{DBLP:journals/tip/ZhangLZZZ15}
Ruimao Zhang, Liang Lin, Rui Zhang, Wangmeng Zuo, and Lei Zhang.
\newblock Bit-scalable deep hashing with regularized similarity learning for
  image retrieval and person re-identification.
\newblock {\em {IEEE} Transactions on Image Processing}, 24(12):4766--4779,
  2015.

\bibitem[\protect\citeauthoryear{Zhao \bgroup \em et al.\egroup
  }{2015a}]{DBLP:conf/cvpr/ZhaoHWT15}
Fang Zhao, Yongzhen Huang, Liang Wang, and Tieniu Tan.
\newblock Deep semantic ranking based hashing for multi-label image retrieval.
\newblock In {\em CVPR}, pages 1556--1564, 2015.

\bibitem[\protect\citeauthoryear{Zhao \bgroup \em et al.\egroup
  }{2015b}]{DBLP:journals/spl/ZhaoLZ15}
Xueyi Zhao, Xi~Li, and Zhongfei~(Mark) Zhang.
\newblock Multimedia retrieval via deep learning to rank.
\newblock {\em {IEEE} Signal Processing Letters}, 22(9):1487--1491, 2015.

\end{thebibliography}
}
\end{document}